\documentclass[10pt,twocolumn,letterpaper]{article}

\usepackage[english]{babel}
\usepackage{times}
\usepackage{epsfig}
\usepackage{subfig}
\usepackage{graphicx}
\usepackage{amsmath}
\usepackage{amssymb}
\usepackage{algorithm}
\usepackage{algpseudocode}
\usepackage{colortbl}
\usepackage{multirow}
\usepackage{xcolor}
\usepackage{booktabs}
\usepackage{titling}
\usepackage[T1]{fontenc}
\usepackage{authblk}
\usepackage[pagebackref=true,breaklinks=true,letterpaper=true,colorlinks,bookmarks=false]{hyperref}
\title{Orientation-Guided Contrastive Learning for UAV-View Geo-Localisation}

\author[1]{Fabian Deuser}
\author[1]{Konrad Habel}
\author[2]{Martin Werner}
\author[1]{Norbert Oswald}

\affil[1]{Institute for Distributed Intelligent Systems, University of the Bundeswehr Munich}
\affil[2]{Department of Big Geospatial Data Management, Technische Universität München}

\begin{document}
\date{}
\maketitle
\begin{abstract}

Retrieving relevant multimedia content is one of the main problems in a world that is increasingly data-driven. With the proliferation of drones, high quality aerial footage is now available to a wide audience for the first time. Integrating this footage into applications can enable GPS-less geo-localisation or location correction. 

In this paper, we present an orientation-guided training framework for UAV-view geo-localisation. Through hierarchical localisation orientations of the UAV images are estimated in relation to the satellite imagery. We propose a lightweight prediction module for these pseudo labels which predicts the orientation between the different views based on the contrastive learned embeddings. We experimentally demonstrate that this prediction supports the training and outperforms previous approaches. The extracted pseudo-labels also enable aligned rotation of the satellite image as augmentation to further strengthen the generalisation. During inference, we no longer need this orientation module, which means that no additional computations are required. We achieve state-of-the-art results on both the University-1652 and University-160k datasets. 

\end{abstract}
%%%%%%%%% BODY TEXT
\section{Introduction}
Geo-Locations from images can help in various situations where accurate real-time kinematic GPS (RTK) is either not available, too expensive or the GPS sensor receives noisy signals~\cite{Chebrolu2019robotlocal}. Especially in cities, the urban canyon effect can create deviations of several metres~\cite{brosh2019accurate}. An alternative to satellite-based radio navigation is the so-called cross-view image-localisation~\cite{lin2013cross}, where the relationship between ground and aerial imagery is utilised for localisation. The current standard datasets like CVUSA~\cite{workmann2015cvusa}, CVACT~\cite{liu2019cvact} and VIGOR~\cite{zhu2021vigor} only feature satellite imagery in combination with street-level images to locate the position of the street-view. 
\begin{figure}[t]
\begin{center}
     \includegraphics[width=1.0\linewidth]{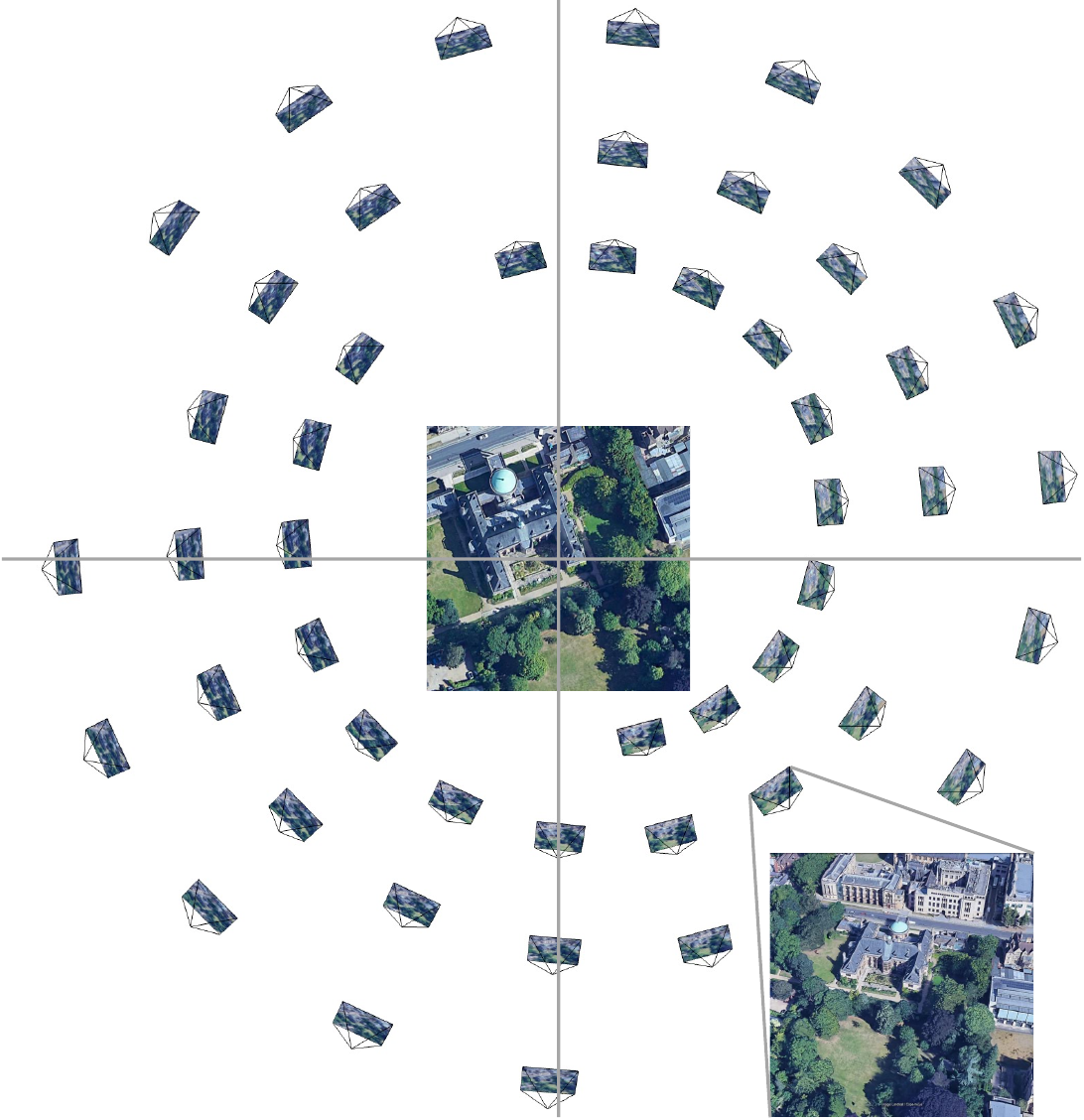}
\end{center}
      \caption{Visualisation of our orientation estimation and division into orientations as pseudo-labels to strengthen the alignment between drones and satellite views.}\label{fig:method}
\end{figure}
Since the proliferation of unmanned aerial vehicles (UAV), so called drones, more and more videos and images from a UAV-perspective are available. Localising these offline as well as online can help in navigation, delivery or large-scale 3D reconstruction tasks~\cite{shi2022beyond, Dissanayaka2022reviewUAV,Sorbelli2023TruckDrone, turki2022mega, Tancik_2022_CVPR}. To develop approaches suitable for UAV-views Zheng et al. introduced the University-1652 dataset~\cite{zheng2020university} to the research community. In this dataset, drone images of buildings have to be matched with the corresponding satellite images. To test the additional generalisation of the approaches, an extension of the dataset, namely University-160k, is released. This extension features additional satellite images of non related regions. These images serve as distraction patterns to increase the level of difficulty and test the transferability in applications. Due to the large satellite database, the efficiency of matching between embeddings must also be taken into account to enable production-ready applications. 

In modern deep learning based approaches~\cite{deuser2023sample4geo,wang2021each,shi2019safa,zhu2022transgeo,zhang2022cross,liu2019lending}, an embedding of both views is learned. The outputed embeddings are then matched based on the cosine similarity or $\ell_2$ distance. 

As previous datasets and approaches~\cite{shi2020looking,zhu2021revisiting,zhu2022transgeo} have shown alignment between the geographical direction of ground and aerial images provides a decisive advantage in prediction. But in the University-160k dataset no orientation information is available. Therefore, we use a 6-DoF localisation technique~\cite{sarlin2019coarse} to estimate the orientations of the individual drone views. These orientations can then be divided into directions to be predicted as labels as shown in Figure~\ref{fig:pipeline}. In this way, the image encoder learns implicitly from which direction the drone image was captured and aligns the outputted embeddings. Additionally we compare between CNN and Transformer-based architectures to explore the best trade-off between efficiency and effectiveness and propose a simple label cleaning in the presence of uncertain views and occlusions.

\section{Related Work}
%several sentence from the related work are commented out since they dont add particular value to the paper, and i need more space
The University-1652~\cite{zheng2020university} dataset is the first multi-view, multi-source benchmark for geo-localisation with drone imagery. The train and test contains 701 unique buildings each from universities with 54 drone images and one satellite image per building. To test the generalisation, it was ensured that there was no overlap between the universities in the train and test split. Zheng et al. used an multi-branch CNN to extract embeddings of the views and optimise it based on an instance loss. During inference the classification layer is not further used.

The drone images in the dataset are synthetic images created using Google Earth Studio. They are created by an automatic render pipeline, which is why the colours are different compared to the satellite images. To fix this, Hu et al.~\cite{hu2020multi} proposed a style alignment based on the colour statistics and a rotation of the images. Since buildings are the focus of the images, Lu et al.~\cite{Lu_2022_ACCV} suggested an context-aware hierarchical representation selection module that learns to focus on feature map regions with high activation's. 

However, Wang et al.~\cite{wang2021each} argue that all information, including background information, can be important because, for example, roads and trees are visible in both the satellite image and the drone image. Therefore, they proposed a pooling based on a ring pattern instead of the normal average pooling at the end of the network. For each image, the feature map is divided into four smaller rings and each ring is pooled separately. The output feature vectors are then used individually for the loss calculation. 

Due to the different viewpoints, an alignment of the feature maps can help to find similarities in the images to be matched. Ming et al.~\cite{ming2022featureseg} proposed a lernable feature segmentation and region alignment based on the outputed heatmap. In their approach pseudo labels corresponding the heatmap are generated and used to classify foreground and background to further support alignment between the different views.

To leverage additional semantic information about the flight height and camera angle Zhu et al.~\cite{zhu2023uavstatus} encode textual features with a BERT model~\cite{devlin2018bert}. In a subsequent fusion module the output of the Transformer is fused with image features outputed by a ResNetV2.

As the cross-view geo-localisation task can be modelled as both classification and retrieval to learn appropriate representations, contrastive loss functions are used. In particular, contrastive loss functions benefit from hard negatives and the sampling of these has been studied by Deuser et al.~\cite{deuser2023sample4geo}. In their work they proposed a location-based sampling method to use hard negatives from the beginning of training. 
\section{Methodology}
As previous work has shown, orientation plays a fundamental role in training~\cite{shi2020looking,zhu2021revisiting,zhu2022transgeo}. For the previous used datasets, the geographical orientation is known a priori when the 360-degree images are extracted. Thus, for example, the rotation of the satellite images can be used as a data augmentation technique in which a previously aligned 360 degree image is then rolled by the amount of rotation. In the University-1652 dataset this orientation is unknown. However, with multiple images of the same object available, the orientation and estimated position in space can be determined from the 3D reconstruction using Structure-from-Motion (SfM) methods~\cite{Schonberger_2016_CVPR}. 
\subsection{Hierarchical Localisation}
\label{sec:hloc}
\begin{figure*}
\begin{center}
     \includegraphics[width=1.0\textwidth]{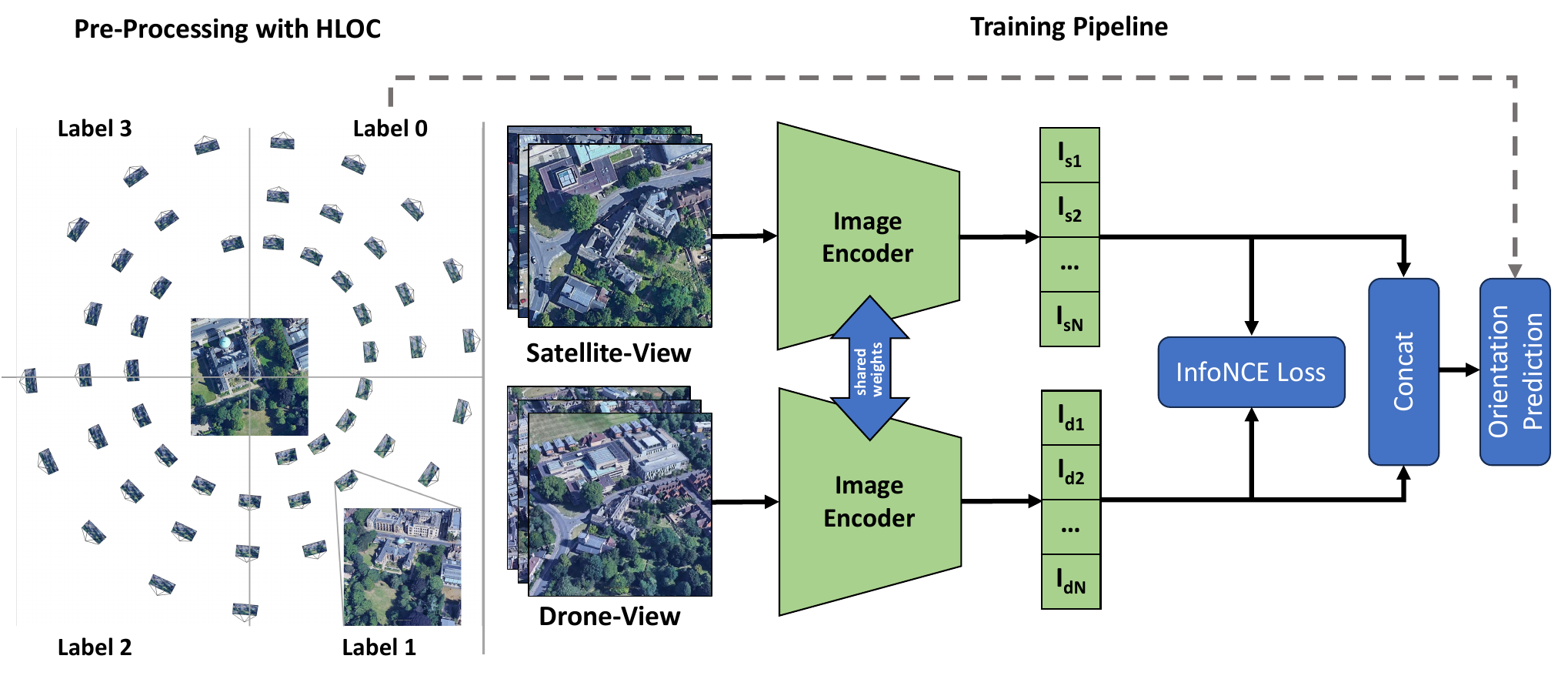}
\end{center}
      \caption{Visualisation of our training pipeline. First we estimate the orientation for the whole training set. Then the extracted angles between satellite and drone-views are binned and used as pseudo-labels during the contrastive training.}\label{fig:pipeline}
\end{figure*}
In the first stage of our approach, the orientations of the UAV views have to be estimated. For this we use an off-the-shelf hierarchical localisation technique (HLOC) by Sarlin et al.~\cite{sarlin2019coarse}. First, global descriptors are generated using a pre-trained CNN. The images are then ordered at a coarse level with k-nearest neighbours on the global descriptors. These so-called prior frames represent location candidates for a scene and are then clustered into places based on their 3D structure which is generated using SfM. Since global features are usually too coarse, local features are also extracted with SuperPoint~\cite{detone2018superpoint}. Between the images in a cluster local features are then matched and the 6-DoF pose is estimated with PnP~\cite{kneip2011pnp}. Due to outliers, RANSAC is used to increase the robustness of the estimation. For predicting the orientation we use the satellite image as anchor and assign each UAV-view a pseudo label based on the angle of the 3D coordinates.

As shown on the left side in Figure~\ref{fig:pipeline} the estimation of the transformation matrices reconstructs the camera path that was taken for the photos. To generate pseudo alignment labels we set the coordinate of the satellite image as our origin and then calculate the angles between each drone view and the sat view. The estimated angles are then used to divide the images into $b$ bins every $x$ degree. We also extract only the angles to regress them, but this can be noisy because the hierarchical localisation is only an estimate. During training the embeddings of satellite and drone view are used to predict the pseudo-labels. During inference these weights are no longer needed. 

To help the model learn the alignment, we replace the random rotation as data augmentation. In 30 \% of the cases the satellite image is rotated and the pseudo label which indicates the orientation to the satellite image is adjusted accordingly as aligned rotation data augmentation.

\subsection{Network Architecture and Loss Function}
Our overall architecture for the metric learning is adapted from Deuser et al.~\cite{deuser2023sample4geo}. Therefore, we use an Siamese network~\cite{koch2015siamese} with a weight-shared pre-trained image encoder~\cite{liu2022convnet,oquab2023dinov2} as depicted in Figure~\ref{fig:pipeline}. In our experiments we compare a Vision Transformer with a state-of-the-art ConvNet, since Deuser et al.~\cite{deuser2023sample4geo} showed advantages of CNNs in geo-localisation with ground-views. As contrastive loss function for the cross-view matching we use the InfoNCE loss~\cite{oord2018representation,he2020momentum}. This loss function contrasts each positive sample against all other negatives sample within a batch. Having two or more drone views of the same building in a batch behaves like label noise for the loss function. To prevent this behaviour we sample only one drone view of a building into a batch. In addition to before described aligned rotation, we use colour jitter, coarse dropout, Gaussian blur and sharpening of the images for data augmentation. During the training phase, we include a lightweight linear layer to predict the generated orientation labels based on the concatenated output features of the satellite and UAV view branches. The lightweight linear layer is not used for inference, thus our method does not require any additional computing power compared to Deuser et al.~\cite{deuser2023sample4geo}. We use AdamW~\cite{loshchilov2017decoupled} as optimiser and use a cosine learning rate scheduler with warmup. As warmup proportion we set 10 \% of the training steps and set the peak maximum learning rate to 4e-5. Label smoothing of 0.1 is used in the InfoNCE loss and in the cross-entropy loss for the discrete orientation prediction. In our experiments we also tried to regress the orientation, therefore we use the mean-squared error (MSE) as loss function. The contrastive and the orientation loss are weighted at a ratio of 2 to 1. We train our approach for one epoch on the University-1652 train set with 8 Nvidia V100 32Gb. 
\subsection{Orientation-Guided Noise Reduction}

Cities are crowded and occlusions of important features and objects are not uncommon. This can also be a problem for the University-160k dataset, as it is almost impossible to locate the building from some images. Additionally it can introduce noise during training and inference as well. Conveniently, the orientation estimation fails for such samples, so we can automatically identify these non-useful samples in our training set. During the training, for both the orientation prediction and the positive part of the contrastive loss, we mask the identified noisy examples. However, since the InfoNCE loss benefits from more negative examples, we keep them as negative examples in the loss function.

\section{Experiments}
In the evaluation of our approach, we first tested on the University-1652 to establish comparability with previous approaches. As can be seen in Table~\ref{tab:uniS2D}, we achieve state-of-the-art results. However, Zhedong et al. presented a challenging version in their new iteration of the University dataset~\cite{zheng2023UVA}. This extension features a total of 160k satellite images during the test period instead of the previous 951. This makes the task much more challenging as the training data is not increased. Thus, this dataset is very well suited for testing the generalisation ability of a model. 

\begin{table}[h]
    \centering
   \begin{center}
    \resizebox{\columnwidth}{!}{ 
    \begin{tabular}{l|cc|cc} \hline \hline
        Approach &\multicolumn{2}{c|}{Drone2Sat}& \multicolumn{2}{c}{Sat2Drone} \\
         & R@1 & AP & R@1  & AP   \\ \hline \hline
        LPN~\cite{wang2021each} & 75.93 & 79.14 & 86.45 & 74.79\\
        DWDR~\cite{Wang2022LearningCG} & 86.41 & 88.41 & 91.30 & 86.02\\
        MBF~\cite{zhu2023uavworth} & 89.05 & 90.61 & 92.15 & 84.45 \\
        Sample4Geo~\cite{deuser2023sample4geo} & 92.65 & 93.81 & 95.14 & 91.39\\ 
        Ours (ConvNext) & 94.72 & 95.63 & 95.57 &93.47\\ 
        Ours (DINOv2) & \textbf{96.13} & \textbf{96.88} & \textbf{96.71} & \textbf{95.25}\\ \hline \hline
    \end{tabular}
    }
   \end{center}
    \caption{\textbf{Quantitative comparison between our approach and state-of-the-art approaches on University-1652~\cite{zheng2020university}.}}
    \label{tab:uniS2D}
\end{table} % sat2drone drone2sat scores erzeugen!

As can be observed in Table~\ref{tab:uniS2D} compared to Table~\ref{tab:uni160k} the baseline LPN suffers from degradation of 75.93\% to 64.85\% in Recall@1 and 79.14 \% to 67.69 \% in average precision. When comparing our results on the University-1652 and University-160k dataset we only observe a minimal drop in performance. We compared the ConvNeXt-XXLarge with a Vision Transformer (ViT) L/14. The ConvNeXt is pre-trained on the LAION dataset~\cite{schuhmann2022laionb} and the Vision Transformer on the LVD-142M dataset~\cite{oquab2023dinov2}. Deuser et al.~\cite{deuser2023sample4geo} showed the advantages of CNNs on varying image size like in 360 Degree views and satellite images. Based on our result this could not be observed for the drone views used in the University-1652 dataset. During the experiments an image size of $384 \times 384$ is used for both, satellite and drone views. The ensemble in Table~\ref{tab:uni160k} is the merged predictions of our ConvNeXt and DINOv2 results.

\begin{table}[h]
    \centering
   \begin{center}
    \resizebox{\columnwidth}{!}{ 
    \begin{tabular}{l|ccccc} \hline \hline
       Approach & R@1 & R@5 & R@10  & AP \\ \hline
        Baseline (LPN)~\cite{zheng2023UVA} & 64.85 & - & - &  67.69\\
        Ours (ConvNeXt) & 93.50 & 97.67 & 98.08  & - \\
        Ours (DINOv2) & 94.60 & 98.05 & 98.38  & -\\
        Ours (Ensemble) & \textbf{95.71} & \textbf{98.69} & \textbf{98.91} & -\\ \hline \hline
    \end{tabular}
    }
    \end{center}
    \caption{\textbf{Quantitative comparison between our approach and the baseline on the extended University-160k dataset.}}
    \label{tab:uni160k}
\end{table}

\section{Ablation Study}
In the ablation study, we investigate the effects of orientation prediction and the influence of coarse-to-fine granular orientations as labels. In addition, different embedding sizes and the effect on performance are tested. 
Because there is always a trade-off between efficiency and effectiveness, we also tested the impact of lower embedding size to enable more efficient search. We compared different pre-trained Vision Transformer~\cite{oquab2023dinov2, ilharco_gabriel_2021_5143773, radford2021learning} and found that the DINOv2~\cite{oquab2023dinov2} provided the best initialisation for the task.

\subsection{Impact of Orientation Directions}
\begin{table}
    \centering
   \begin{center}
    \resizebox{\columnwidth}{!}{ 
    \begin{tabular}{l|ccccc} \hline \hline
       Orientation & R@1 & R@5 & R@10  & AP \\ \hline
        None & 95.22 & 98.69 & 99.15  & 96.04\\
        b=4 & 95.71 & 98.93 & 99.31 &  96.47\\
        b=8 & 95.90 & 98.95 & 99.28  & 96.63\\
        b=16 & 95.50 & 98.91 & 99.34  & 96.29\\
        b=32 & \textbf{96.13} & \textbf{99.30} & \textbf{99.50}  & \textbf{96.88}\\
     \hline \hline
    \end{tabular}
    }
    \end{center}
    \caption{\textbf{Quantitative comparison different orientation labels on the University-1652 dataset.}}
    \label{tab:orientationablation}
\end{table}
As described in Section~\ref{sec:hloc} 3D-coordinates are generated for each image of the train set. The angles between the satellite image and the corresponding drone images are calculated. However, errors can occur because no ground truth is known and the coordinates are only estimated by HLOC~\cite{sarlin2019coarse}. Therefore, we tested different abstractions and discretized the angles as categorical labels. For this purpose, we have divided the image into 4, 8, 16, 32 bins, similar to what can be observed in Figure~\ref{fig:pipeline}, in order to use these as pseudo class-labels.
Table~\ref{tab:orientationablation} shows the results using no orientation prediction (None) and orientation prediction with different bin sizes $b$. The orientation prediction is formulated as classification task on a limited number of discrete bins $b$. We also tested the formulation as a regression task using MSE-Loss as training objective without discrete bins, but this does not leads to better results. For the University-160k challenge we used $b=32$ during training.

\begin{table}
    \centering
   \begin{center}
    \resizebox{\columnwidth}{!}{ 
    \begin{tabular}{ll|ccccc} \hline \hline
       Model & Dim & R@1 & R@5 & R@10  & AP \\ \hline
        ConvNeXt~\cite{deuser2023sample4geo}  & 1024 & 92.56 & - & -  & 93.81\\
        ViT-S/14  & 384 & 92.59 & 97.79 & 98.58  & 93.79\\
        ViT-B/14  & 768 & 94.77 & 98.71 & 99.15 &  95.67\\
        ViT-L/14  & 1024 & \textbf{96.13} & \textbf{99.30} & \textbf{99.50}  & \textbf{96.88}\\
     \hline \hline
    \end{tabular}
    }
    \end{center}
    \caption{\textbf{Quantitative comparison of different Transformer Sizes on the University-1652 dataset.}}
    \label{tab:embsizeablation}
\end{table}
\subsection{Embedding Size}
Finding locations fast and efficient is a key issue that image-based location estimation must address in application. In the literature, neighbourhood graphs are constructed or the vectors are quantised~\cite{iwasaki2018optimization, guo2020accelerating} to enable a more efficient search. Another very simple method to minimise the computational effort for the search is to reduce the embedding size. As shown in table~\ref{tab:embsizeablation} we can easily scale down our ViT-based approach and achieve competitive results even with the smallest embedding size of 384 compared to the state-of-the-art~\cite{deuser2023sample4geo} which uses a ConvNeXt-base with an embedding size of 1024. 

\section{Conclusion}
In our paper we present an easy-to-use approach to generate additional orientation labels for drone views.  These pseudo-labels show a positive impact on the results despite the lack of ground truth label, as we demonstrate experimentally. Our solution does not add any additional computational cost for inference and can be scaled down to a certain extent in terms of embedding size and model size. With our approach, we achieve new state-of-the-art performance on the University-1652 dataset and also demonstrates high generalisation abilities on the University-160k dataset. This allows us to take a step forward in bringing drone-view geo-localisation into application.

\section{Discussion}
During our work for this paper we discovered two challenges that also facilitate future research directions. First, the given drone images in the University-1652 dataset are generated with Google Earth Studio. Thus, they are synthetic in nature and additional dynamics such as lighting and weather effects are missing. To test models for their generalisation ability in more challenging environments and situations, new datasets with varying dynamics are needed. 

Secondly, in both the training and validation splits, buildings are very close to each other. Drone views can capture two buildings and it is not possible to assign exactly to which building it belongs. This results from the setting to treat drone images individually and not as a sequence. This behaviour can also be seen in the R@5 score in Table~\ref{tab:uni160k}.

\section{Acknowledgement}
The authors gratefully acknowledge the computing time granted by the Institute for Distributed Intelligent Systems and provided on the GPU cluster Monacum One at the University of the Bundeswehr Munich.
{\small
\bibliographystyle{ieeetr}
\bibliography{egbib}
}

\end{document}